# GEN: Highly Efficient SMILES Explorer Using Autodidactic Generative Examination Networks

Ruud van Deursen [a], Peter Ertl [b], Igor V. Tetko [c], Guillaume Godin [a]

[a] Firmenich SA, Research and Development, Rue des Jeunes 1, CH-1227 Les Acacias, Geneva, Switzerland

[b] Novartis Institutes for BioMedical Research, Novartis Campus, CH-4056 Basel, Switzerland

[c] BIGCHEM GmbH and Institute of Structural Biology, Helmholtz Zentrum München - German Research Center for Environmental Health (GmbH), Ingolstädter Landstraße 1, D-85764 Neuherberg, Germany

**Abstract**

Recurrent neural networks have been widely used to generate millions of *de novo* molecules in a known chemical space. These deep generative models are typically setup with LSTM or GRU units and trained with canonical SMILES. In this study, we introduce a new robust architecture, Generative Examination Network GEN, based on bidirectional RNNs with concatenated sub-models to learn and generate molecular SMILES within a trained target space. GENs autonomously learn the target space in a few epochs while being subjected to an independent online examination to measure the quality of the generated set. Here we have used online statistical quality control (SQC) on the percentage of valid molecular SMILES as examination measure to select the earliest available stable model weights. Very high levels of valid SMILES (95-98%) can be generated using multiple parallel encoding layers in combination with SMILES augmentation using unrestricted SMILES randomization. Our architecture combines an excellent novelty rate (85-90%) while generating SMILES with strong conservation of the property space (95-99%). Our flexible examination mechanism is open to other quality criteria.

**Keywords**

Autonomous learning, GEN, GAN, RNN, LSTM, GRU, biLSTM, biGRU, AI, SMILES, generator.

**Introduction**

Exploration of chemical space for the discovery of new molecules is a key challenge for the chemical community, e.g. pharmaceutical and chemical entities [1,2]. Previously, exhaustive enumeration has been introduced with the creation of 26M, 1G and 1.7G molecules in the databases GDB11, GDB13 and GDB17, respectively [3]. Exhaustive enumeration critically depends on knowledge rules specified by a chemist to restrict the combinatorial explosion of possible molecules. Consequently, exhaustive enumeration may generate a realistic but more biased chemical space. More recently, AI methods have been emerging rapidly and have proven successful for text learning [4] and application in drug discovery [5]. Deep generative models based on the SMILES syntax have been proven as highly successful [6]. A recent publication shows that the architecture with a classical recurrent network introduces a bias to the generated space. These results have been confirmed in a recently published work on GDB13, showing that at most 68% of GDB-13 was reproduced using a deep generative model to reproduce the space [3]. SMILES [7] is a very simple text representation of molecules. It is "readable" by chemists and is quickly translated into molecules with RDKit or other cheminformatics toolkits. Other 1D string encoders like InCHI [8] or DeepSmiles [9] were found to be not better than SMILES to generate molecules [3, 10]. Since 2016, SMILES-based machine-learned methods are used to produce *de novo* molecules. These methods include Variational AutoEncoders (VAEs), Recurrent Neural Network (RNN), Generative Adversarial Networks (GANs) and reinforcement learning (RL) [11]. Another alternative to build generative molecules models without SMILES is based on molecular graph representation [11]. Contrary to earlier reports [11], we demonstrate herein that text learning on SMILES is highly efficient to explore the training space with a high degree of novelty. Herein we have modified a previously reported algorithm [6] and use bidirectional RNN layers for better generation results. The neural network of the generator is subsequently converted into a generative examination network (GEN). In GENs, the generative models autonomously learn chemical rules and are free to extract and generalize their rules to reconstruct the training set without being subjected to expert constraints. During training of the models, the learning progress of the generators is periodically assessed using an independent online examination mechanism without direct feedback to the student. In this GEN we use an online generator that applies a statistical quality control after every training epoch, measuring the percentage of validity for a statistical set of generated SMILES. This mechanism is an early stopping function and prevents the network from overfitting the training set to keep the highest degree of generativity. In GENs, the generator and examination methods are open to any other generative network and examination methods, including simple metrics or more advanced models. Our calculations based on the publicly available dataset PubChem [12], clearly demonstrate that the use of bidirectional layers systematically improves the capability of the GEN to generate a vast set of new SMILES within the property space of the training set. Following excellent results of SMILES augmentation for smaller datasets to predict physico-chemical properties [13] and generators [14], we have used SMILES augmentation as an approach to increase the size of the training set and to enhance the diversity of SMILES presented to our GENs.

**Methods**

**Preparation of datasets and encoding.** PubChem database was downloaded in March 2019 as SDF. The canonical SMILES string *PUBCHEM_OPENEYE_CAN_SMILES* was extracted, split into fragments and converted into canonical SMILES using RDKit version 2019.03.3 [15,16]. Only organic molecules, i.e. those that contain at least one carbon and all other atoms are part of the subset H, B, C, N, O, F, S, Cl, Br or I, were retained and then deduplicated to produce a set of unique SMILES. From this dataset, we extracted a representative set of 225k fragment-sized molecules typically explored in the pharmaceutical and olfactive industries [6,17]. Prior to training, the SMILES were either converted to the canonical form or augmented as detailed in the results. Double character atoms were replaced by single characters. The characters Cl, Br and [nH] were modified to L, R and A, respectively. Stereochemistry was removed, replacing [C@H], [C@@H], [C@@] and [C@] by C as well as removing the characters / and \ used for double bond stereochemistry. The molecules were tokenized by making an inventory of observed characters followed by decoding the molecules. The generated text corpus was converted to a training set pairing the next available characters (labels) to the previously observed sentence, which are presented as one-hot encoded feature matrices to the network.

**Architecture.** Modelling was performed using the open source libraries Tensorflow [18] and Keras [19]. The method was programmed in Python and code is freely available under a clause-3 BSD license [20,21]. Architectures used for GENs were composed of an embedding biLSTM- or LSTM-layer, followed by a second encoding biLSTM- or LSTM-layer, a dropout layer (0.3) and a dense layer to predict the next character in the sequence (Figure 1). For Architecture A and B, we also tested biGRU and GRU-layers for embedding and encoding. For consistency of the architecture, LSTM and GRU units were not mixed. Several runs were evaluated to reduce the set of hyperparameters. Here we have evaluated LSTM and GRU units with layer sizes of 64 and 256. The Dense layer had a size equal to the number of unique characters observed in the training set. Architectures C and D with multiple parallel encoding layers were evaluated using merging by concatenation, averaging or learnable weighted average (Figure 1). The code for the learnable weighted average can be downloaded [21].

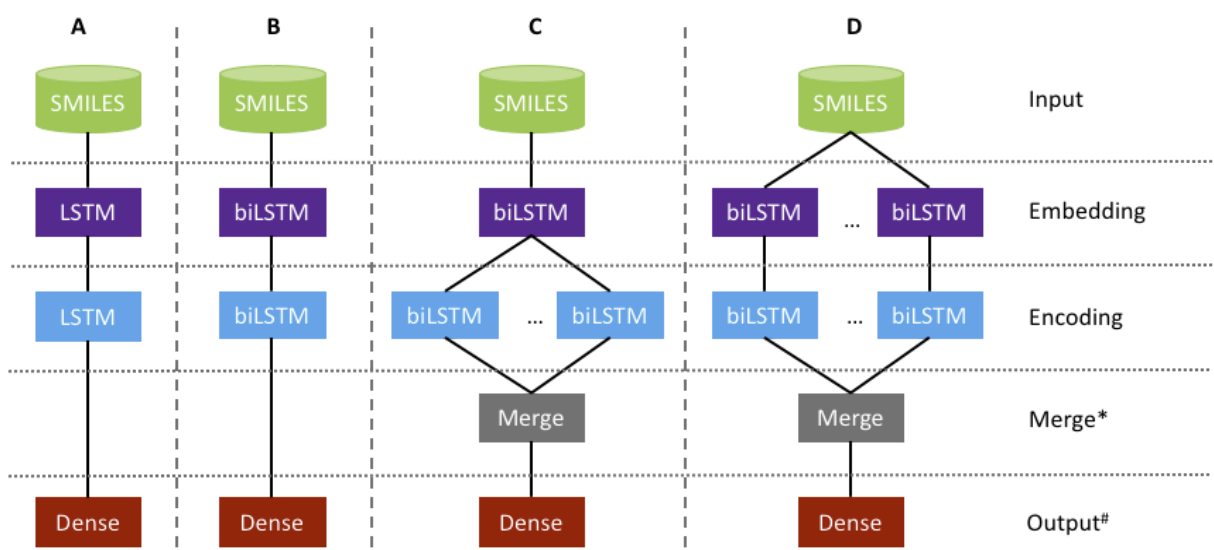

*Figure 1: Tested architecture for SMILES generation.* Architecture with two consecutive biLSTM layers used for deep-generative models for SMILES generation  A) Original architecture with two consecutive LSTM layers, followed by a Dense output layer to predict the next character. B) Modification architecture with two consecutive bidirectional LSTM layers. C) Advanced architecture with one embedding biLSTM layers followed by multiple parallel bidirectional encoding layers and a merging layer (concatenated, averaged or learnable average). D) Advanced architecture using parallel concatenated architectures with multiplication of embedding and encoding layers. These layers are merge by concatenation, averaging or learnable weighted average.

**Training of architecture with online statistical quality control.**

It is widely known that LSTM is based on conservative long-range memory. Architectures A and B produced mostly canonical SMILES (>92%) when trained with a set of canonical SMILES [16,22]. In order to improve the explorative nature of the GENs, we used a set of randomized SMILES. Early stopping was used to avoid overfitting and memorization of the training set [23]. In neural networks based on Keras, early stopping is applied using Callback functions (keras.callbacks.Callbacks). In our workflow (figure 2), we have modified the existing EarlyStopping function, which now performs an online statistical quality control (SQC) on the percentage of valid generated SMILES strings after every epoch [24]. On the training start, the Callback function was parameterized with a target percentage, e.g. 97%, along with the sample size (**$N_{sample}$** = 300). Optionally, the size of the population **($N_{pop}$)** was specified in the callback method. If no value was specified, the size of the population is assumed to be very large i.e. **$N_{pop}$** >> **$N_{sample}$**. Based on the specified parameters, the upper and lower margin were computed using a 95% confidence interval (CI) [25]. The callback function stopped training early, if the trained model showed stable generation results within the 95% CI for 10 consecutive epochs to exclude incidental good results (or as specified by the user, modifying the argument *patience*). The EarlyStopping counter was reset if the percentage of valid structures fell below the lower margin of the computed stability interval. Upon completion of training, the earliest available model was selected and used to generate 2k SMILES strings for evaluation. Evaluations on quality were performed using three independently trained models and their average value and standard mean error are reported for all results. To have an objective assessment of quality, SMILES with easy-detectable errors, i.e. a mismatch for ring and branch closure characters, were included in the evaluation.

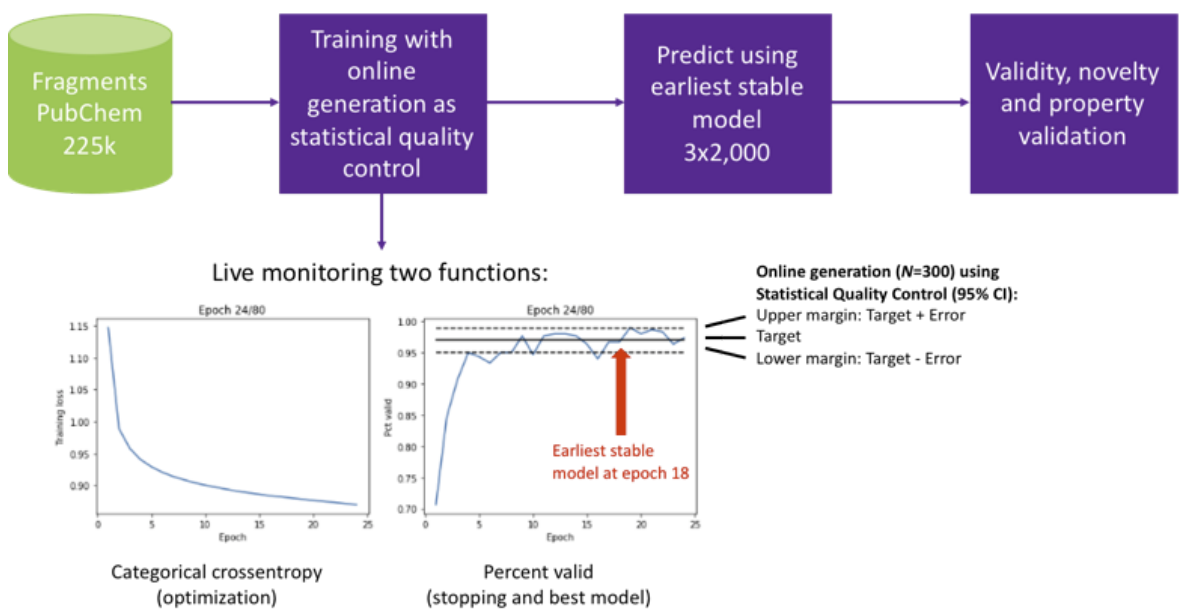

*Figure 2: Modeling workflow used for every architecture/hyperparameter search. The autodidactic generator models learn independently a probability for the next logical character. At every epoch, i.e. online the generator generates a statistical sample of 300 SMILES Strings with are examined using statistical quality control as examination criteria. Upon completion of the training, the earliest stable model that satisfies the quality criteria is selected and evaluated based on a generated sample.*

**Percentage of valid molecules, uniqueness and training compounds.** For all generated molecules, the molecules were considered as valid if they were successfully converted to an InChIKey using RDKit [15]. The percentage of valid molecules, Validity %, was measured as a ratio of the number of valid molecules to the number of generated SMILES. All valid molecules were reduced to unique compounds. The uniqueness, Uniqueness %, was expressed as a number of unique molecules divided by the number of valid molecules. The percentage of known training compound, Training %, was computing dividing the number of generated SMILES known in the training set, by the number of valid SMILES.

**Property distributions and percentage match.** For the model with architecture C (biLSTM-biLSTM 256/256 with four concatenated encoding layers), 200 sets of 10k molecules were generated to create 2 M molecules. For three sets of 10k molecules at the beginning (early) and end (late) of generation process, we have calculated property distributions and compared to the property distributions of the training set. Four classes of properties have been evaluated: A size comparison was performed using SMILES length (measured as number of characters), heavy atom count (HAC, counting all non-hydrogen atoms) and molecular weight; Polarity was evaluated using LogP and TPSA; Topological properties were compared using the number of rotatable bonds, fraction of cyclic, conjugated or aromatic atoms; A comparison on elemental composition was performed based on fractions of carbon, nitrogen or oxygen atoms in the molecules. All distributions are displayed in figure 3 and the percentage match for the distributions of the generated space **A** and training space **B** was computed using the continuous Tanimoto coefficient **T(A,B)** equation 1 [26] and is summarized in table 5.

**Equation 1:**

$$T(A,B) = \frac{\sum_i A_i B_i}{\sum_i A_i^2 + \sum_i B_i^2 - \sum_i A_i B_i} \times 100\%$$

The Jensen-Shannon divergence **JSD(A,B)** between the normalized distributions **A** and **B** was computed applying equation 2:[27,28]

**Equation 2:**

$$JSD(A,B) = H\left(\sum_{d \in \{A,B\}} a_i d_i\right) - \left(\sum_{d \in \{A,B\}} a_i H d_i\right)$$

**Novelty analysis** For architectures A (LSTM-LSTM 256/256), B (biLSTM-biLSTM 256/256) and C (biLSTM-biLSTM 256/256 with 4 concatenated encoding layers), 200 sets of 10k molecules were generated to create a total of 2M SMILES strings for each model. Every set of 10k molecules was considered a time point *t* in the analysis. For every set, all molecules were compared against all previously generated molecules and considered new if the molecule was identified by a new unique InChIKey. The percentage of new molecules was subsequently expressed as number of new molecules divided by the number of valid molecules (Figure 4A). All unique molecules were summed over time (figure 4B). An overall percentage of efficiency was expressed as number of valid unique molecules divided by the number of generated SMILES strings (figure 4B). The novelty results were subsequently split by HAC to perform a novelty and saturation analysis by molecular size in the range of 4-24 atoms. (Figure 5).

**Size analysis of the training set.** The impact of the size of the training set on the GENs was evaluated using random fragments subsets from PubChem [12], Zinc15 [29] and Chembl24 [30]. We have evaluated the sizes 225k, 45k and 9k with a random SMILES augmentation of

5 attempts SMILES per molecule. The augmented smiles were deduplicated and the number of real augmentations may vary for each dataset (table 6). The datasets for Chembl and Zinc were subjected to the data preparation as described earlier for PubChem. The datasets were evaluated using architecture C with a biLSTM embedding layer of size 128 and 4 concatenated parallel biLSTM encoding layers of size 64. The datasets for Chembl and Zinc are available with the download of the source code [21].

**Results**

We evaluated the performance of LSTM and GRU layers on architectures A and B (table 1). For both architectures, the use of LSTM units led to higher percentages of valid SMILES strings and generated a very high percentage of valid SMILES (97%). All other computed metrics, i.e. percentage uniqueness and percentage training compounds, showed only minor fluctuations between the tested architectures. The use of GRU and biGRU layers resulted in inferior results and were discontinued in this study.

**Table 1: Comparison architecture A and B and comparing LSTM to GRU.**

| Architecture | Layer size | Best Model Epoch # | Validity % | Uniqueness % | Training % | Length match %[1] | HAC match %[2] |
|---|---|---|---|---|---|---|---|
| A: LSTM-LSTM | 256 / 256 | 12, 17, 20 | **96.7 +/- 0.4** | 99.9 +/- 0.1 | 15.0 +/- 0.7 | 98.2 +/- 0.9 | 94.0 +/- 1.8 |
| A: GRU-GRU | 256 / 256 | 15, 15, 15 | 91.8 +/- 0.7 | 99.9 +/- 0.1 | 12.6 +/- 0.8 | 98.3 +/- 0.4 | 94.6 +/- 1.3 |
| B: biLSTM-biLSTM | 256 / 256 | 6, 7, 10 | **97.1 +/- 0.4** | 99.9 +/- 0.1 | 13.1 +/- 0.5 | 98.2 +/- 0.6 | 93.9 +/- 0.8 |
| B: biGRU-biGRU | 256 / 256 | 11, 11, 11 | 95.6 +/- 0.6 | 99.9 +/- 0.1 | 15.0 +/- 0.5 | 98.3 +/- 0.3 | 93.1 +/- 1.4 |

1) Length match for SMILES length distributions of the training set and generated set (see methods).
2) HAC match for the atom count distributions of the generated set and training set (see methods).

We extended our analysis to all four architectures A-D, followed by an evaluation using the same quality metrics (Table 2). Several important results were observed. Firstly, the increase of the layer size in architecture A and B led to a lower and more stable number of epochs needed to complete training. The use of larger layer did not significantly improve the generative performance of the model and the property match to the training stage was stable at 98% and 94% for SMILES Length and HAC, respectively. Secondly, parallelization of the encoding layers (architecture C) provided a very good coverage of chemical space by generating a large amount of new molecules. In particular, merging the parallel layers using concatenation significantly improved the performance of the generator pushing the percentage match of the property space to 98.5% and 97.4% for SMILES Length and HAC, respectively. Merging by averaging of a learnable weighted average moderately improved the results of the HAC match by 0.8-1.5% compared to the architecture B.

Thirdly, architecture D with parallel embedding-encoding layers (architecture D) displayed moderately inferior results in comparison with architecture C (HAC -1.9%). However, the result was still moderately better than using architecture B (HAC +1.5%), suggesting the use of multiple parallel encoding layers was beneficial to train a stable generator. These results also suggested that a single bidirectional embedding layer was sufficient to describe the SMILES strings in the training space.

Lastly, we tested two alternative architectures to better understand the importance of the bidirectional nature of the embedding and encoding layers (table 2, last two lines). Modification of the embedding layer from LSTM to biLSTM significantly reduces the number of epochs to obtain a stable generator. Indeed, when comparing the architectures LSTM-LSTM and biLSTM-biLSTM (table 2, last two lines), the number of epochs is decreased by 48% from an average of 23 to 12 epochs. Introduction of bidirectional encoding layers improved the ability of the model to better reproduce the training space (HAC +1.9%). In conclusion, architecture C with a model based on a single bidirectional embedding layer followed by multiple concatenated bidirectional encoding layers provided the best performance (in bold in Table 2).

**Table 2: Comparison architectures A, B, C and D.** Best architecture is highlighted in bold.

| Architecture | Merge mode | Layer count | Layer size | Best Model Epoch # | Validity % | Uniqueness % | Training % | Length match %[1] | HAC match %[2] |
|---|---|---|---|---|---|---|---|---|---|
| A: LSTM-LSTM | - | 1 / 1 | 64 / 64 | 54, 72, 63 | 95.4 +/- 0.4 | 99.9 +/- 0.1 | 12.0 +/- 0.9 | 98.2 +/- 0.3 | 94.0 +/- 0.9 |
| B: biLSTM-biLSTM | - | 1 / 1 | 64 / 64 | 20, 22, 28 | 96.5 +/- 0.5 | 99.9 +/- 0.1 | 12.5 +/- 0.9 | 97.9 +/- 0.5 | 94.9 +/- 0.8 |
| A: LSTM-LSTM | - | 1 / 1 | 256 / 256 | 17, 17, 20 | 96.7 +/- 0.4 | 99.9 +/- 0.1 | 15.0 +/- 0.7 | 98.2 +/- 0.9 | 94.0 +/- 1.8 |
| B: biLSTM-biLSTM | - | 1 / 1 | 256 / 256 | 6, 7, 10 | 97.1 +/- 0.4 | 99.9 +/- 0.1 | 13.1 +/- 0.5 | 98.2 +/ 0.6 | 93.9 +/- 0.8 |
| **C: biLSTM-biLSTM** | **Concatenated** | **1 / 4** | **64 / 64** | **10, 14, 16** | **97.0 +/- 0.3** | **99.9 +/- 0.0** | **11.9 +/- 0.6** | **98.5 +/- 0.3** | **97.4 +/- 0.5** |
| C: biLSTM-biLSTM | Average | 1 / 4 | 64 / 64 | 11, 15, 15 | 97.2 +/- 0.3 | 99.9 +/- 0.1 | 12.5 +/- 0.3 | 98.6 +/- 0.2 | 96.1 +/- 0.7 |
| C: biLSTM-biLSTM | Learnable average | 1 / 4 | 64 / 64 | 15, 17, 23 | 97.6 +/- 0.2 | 99.9 +/- 0.0 | 14.6 +/- 0.2 | 97.4 +/- 0.4 | 94.8 +/- 1.2 |
| D: biLSTM-biLSTM | Concatenated | 4 / 4 | 64 / 64 | 11, 11, 9 | 96.9 +/- 0.3 | 99.9 +/- 0.0 | 14.4 +/- 0.5 | 97.4 +/- 0.2 | 95.6 +/- 1.2 |
| D: biLSTM-biLSTM | Average | 4 / 4 | 64 / 64 | 15, 17, 14 | 96.7 +/- 0.1 | 99.9 +/- 0.0 | 11.9 +/- 0.2 | 98.1 +/- 0.5 | 95.3 +/- 1.1 |
| D: biLSTM-biLSTM | Learnable average | 4 / 4 | 64 / 64 | 12, 25, 18 | 95.6 +/- 0.1 | 99.9 +/- 0.0 | 10.4 +/- 0.5 | 98.0 +/- 0.2 | 96.2 +/- 0.6 |
| **Influence of bidirectionality:** | | | | | | | | | |
| LSTM-LSTM | Concatenated | 1 / 4 | 64 / 64 | 20, 17, 31 | 96.8 +/- 0.4 | 99.9 +/- 0.1 | 13.4 +/- 0.5 | 97.6 +/- 0.8 | 94.8 +/- 1.3 |
| biLSTM-LSTM | Concatenated | 1 / 4 | 64 / 64 | 9, 14, 9 | 97.1 +/- 0.3 | 99.9 +/- 0.1 | 13.2 +/- 0.5 | 97.7 +/- 0.9 | 95.5 +/- 1.4 |

1) Length match for SMILES length distributions of the training set and generated set (see methods).
2) HAC match for the atom count distributions of the generated set and training set (see methods).

We evaluated the influence of the number of parallel concatenated layers in architecture C (see table 3). Their increase reduced the number of epochs required to converge the generator. Simultaneously, it improved the ability of the generator to reproduce the property distribution of the training space. However, after reaching a plateau (with 4 layers) the introduction of new layers overcomplicated networks due to the increasing of the number of hyperparameters, which resulted in performance drop.

**Table 3: Optimal number of parallel encoding layers in Architecture C**

| Architecture | Merge mode | # Layers | Layer sizes | Best model epoch # | Validity % | Uniqueness % | Training % | Length match %[1] | HAC match %[2] |
|---|---|---|---|---|---|---|---|---|---|
| B: biLSTM-biLSTM | - | 1 / 1 | 64 / 64 | 20, 22, 28 | 97.1 +/- 0.4 | 99.9 +/- 0.1 | 13.1 +/- 0.5 | 98.2 +/- 0.6 | 93.9 +/- 0.8 |
| C: biLSTM-biLSTM | Concatenated | 1 / 2 | 64 / 64 | 19, 19, 19 | 97.8 +/- 0.4 | 99.9 +/- 0.1 | 12.5 +/- 0.4 | 97.3 +/- 0.4 | 96.1 +/- 0.1 |
| **C: biLSTM-biLSTM** | **Concatenated** | **1 / 3** | **64 / 64** | **12, 12, 12** | **97.2 +/- 0.2** | **99.9 +/- 0.0** | **12.2 +/- 0.4** | **98.6 +/- 0.3** | **96.9 +/- 0.8** |
| **C: biLSTM-biLSTM** | **Concatenated** | **1 / 4** | **64 / 64** | **10, 14, 16** | **97.0 +/- 0.3** | **99.9 +/- 0.0** | **11.9 +/- 0.6** | **98.5 +/- 0.3** | **97.4 +/- 0.5** |
| C: biLSTM-biLSTM | Concatenated | 1 / 5 | 64 / 64 | 8 | 95.9 +/- 0.3 | 99.9 +/- 0.0 | 13.5 +/- 1.0 | 97.6 +/- 0.2 | 97.2 +/- 0.3 |
| C: biLSTM-biLSTM | Concatenated | 1 / 6 | 64 / 64 | 8 | 95.9 +/- 0.2 | 99.9 +/- 0.1 | 10.1 +/- 0.4 | 96.3 +/- 0.3 | 93.9 +/- 0.7 |
| C: biLSTM-biLSTM | Concatenated | 1 / 7 | 64 / 64 | 7 | 96.8 +/- 0.4 | 99.9 +/- 0.0 | 14.0 +/- 1.0 | 97.6 +/- 0.6 | 95.9 +/- 0.5 |
| C: biLSTM-biLSTM | Concatenated | 1 / 8 | 64 / 64 | 6, 6, 6 | 96.2 +/- 0.7 | 99.9 +/- 0.0 | 13.6 +/- 0.1 | 98.0 +/- 0.7 | 94.8 +/- 0.8 |
| C: biLSTM-biLSTM | Concatenated | 1 / 16 | 64 / 64 | 5, 5, 5 | 95.9 +/- 0.3 | 99.9 +/- 0.0 | 13.5 +/- 1.0 | 96.6 +/- 0.7 | 93.1 +/- 0.7 |

1) Length match for SMILES length distributions of the training set and generated set (see methods).
2) HAC match for the atom count distributions of the generated set and training set (see methods).

Overall, the best results were achieved with architecture C using one biLSTM embedding layer and 4 parallel concatenated biLSTM encoding layers. Using this architecture, an augmentation study was performed. Augmentation was done offline before the training. The new non-canonical SMILES were generated with RDKit by setting option doRandom=True, which was recently successfully used to develop models predictive models for physico-chemical properties [13]. As expected, the augmentation improved the percentage of generated valid SMILES while lowering the number of training epochs. The performed analysis indicated that a 4-fold augmentation provided the optimal result (Table 4). Additional augmentations only moderately improved the capability of the model to better reproduce the property space of the training set.

**Table 4: Augmentation effect on architecture C biLSTM-biLSTM with layer sizes 64/64 and 4 concatenated encoding layers**

| Smiles | Augmentation | Best model epoch # | Validity % | Uniqueness % | Training % | Length match %[1] | HAC match %[2] |
|---|---|---|---|---|---|---|---|
| Canonical 1x | 1x | 9, 9, 7 | 96.6 +/- 0.5 | 99.9 +/- 0.1 | 16.2 +/- 1.5 | 93.3 +/- 0.3 | 92.0 +/- 0.5 |
| Random | 1 | 10, 14, 16 | 97.0 +/- 0.3 | 99.9 +/- 0.0 | 11.9 +/- 0.6 | 98.5 +/- 0.3 | 97.4 +/- 0.5 |
| Random | 2 | 5, 5, 5 | 97.3 +/- 0.1 | 99.9 +/- 0.0 | 13.9 +/- 0.5 | 97.7 +/- 0.4 | 94.5 +/- 0.8 |
| Random | 3 | 4, 6, 4 | 97.9 +/- 0.3 | 99.9 +/- 0.0 | 13.6 +/- 0.5 | 98.8 +/- 0.1 | 96.5 +/- 0.2 |
| **Random** | **4** | **4, 3, 4** | **98.2 +/- 0.4** | **99.9 +/- 0.0** | **11.6 +/- 0.5** | **98.8 +/- 0.3** | **97.1 +/- 0.2** |
| Random | 5 | 4, 4, 4 | 98.3 +/- 0.3 | 99.9 +/- 0.0 | 11.2 +/- 0.5 | 97.3 +/- 0.7 | 96.6 +/- 0.3 |
| *Random* | *10* | *4, 4, 4* | *98.3 +/- 0.3* | *99.9 +/- 0.0* | *14.2 +/- 0.5* | *98.4 +/- 0.4* | *98.2 +/- 0.5* |

1) Length match for SMILES length distributions of the training set and generated set (see methods).
2) HAC match for the atom count distributions of the generated set and training set (see methods).

After selection of best architecture (C / BiLSTM-BiLSTM / 256-256 / 4 concatenated), we generated 2M SMILES strings. We computed 12 molecular properties (see methods) at the beginning of generation (early) and after 2M (late) of generated SMILES. The computed distributions were compared to the distributions of the training space (figure 3). All distributions profiles indicate strong ability of the model to produce new SMILES covering the property space of the training set. As expected, we observed a shift for all distributions correlated to molecule size. This observation suggests that the generator starts to saturate the chemical space of smaller molecules. Indeed, the distributions of molecules with sizes 5 and 6 were close to 0 after generation of 250k and 1M SMILES, respectively. The reduced error observed for the Jensen-Shannon divergence on all distributions suggested that the property distributions of the created SMILES were stable after 2M generated SMILES (table 4).

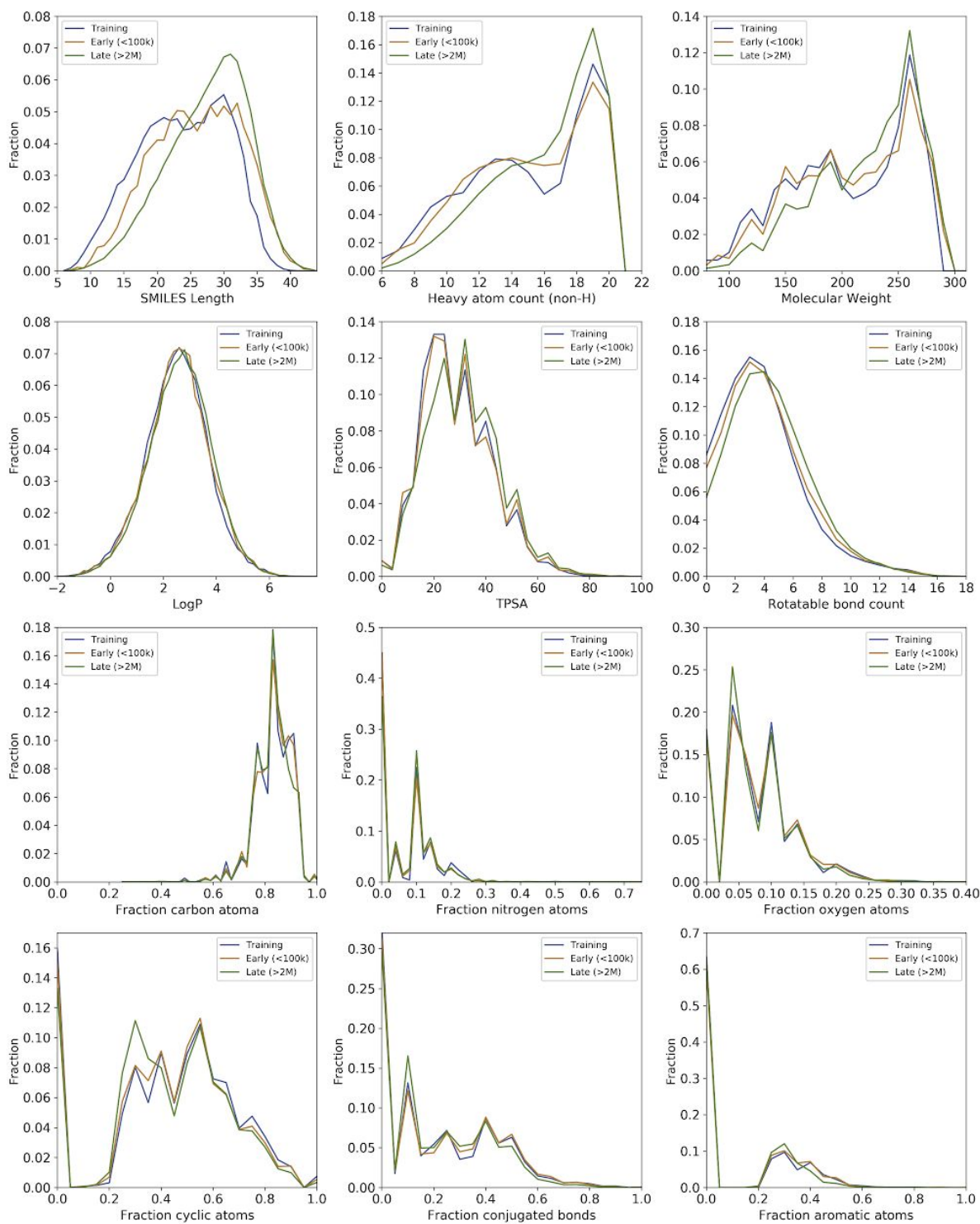

*Figure 3: **Property distributions** for analysed properties including size, topology, polarity and atom compositions. The data are shown for the training set (blue), early generation (orange) and late generation (green). The observed property shifts are due to the saturation for generation of molecules with smaller sizes (see also Figure 4).*

Table 5: Percentage match measured as continuous Tanimoto (eq. 1) between the distributions of the training space and generated compounds at early and late stage generation.

|  | Percentage match (Tanimoto) | | Jensen-Shannon Divergence | |
| --- | --- | --- | --- | --- |
|  | Early stage (10k) | Late stage (2M) | Early stage (10k) | Late stage (2M) |
| **Size:** | | | | |
| SMILES Length | 94.1 +/- 0.4 | 84.6 +/- 0.1 | 0.170 +/- 0.004 | 0.252 +/- 0.000 |
| Heavy atom count (HAC) | 98.8 +/- 0.2 | 94.1 +/- 0.1 | 0.058 +/- 0.004 | 0.142 +/- 0.000 |
| Molecular Weight (MW) | 97.4 +/- 0.2 | 92.7 +/- 0.1 | 0.124 +/- 0.002 | 0.187 +/- 0.000 |
| **Polarity:** | | | | |
| LogP | 99.6 +/- 0.0 | 99.1 +/- 0.0 | 0.042 +/- 0.002 | 0.055 +/- 0.001 |
| TPSA | 99.6 +/- 0.1 | 95.7 +/- 0.1 | 0.044 +/- 0.001 | 0.097 +/- 0.000 |
| **Topology:** | | | | |
| Rotatable bond count | 99.5 +/- 0.1 | 96.5 +/- 0.0 | 0.042 +/- 0.002 | 0.099 +/- 0.001 |
| Fraction cyclic | 99.2 +/- 0.2 | 95.6 +/- 0.1 | 0.051 +/- 0.002 | 0.106 +/- 0.000 |
| Fraction conjugated | 99.6 +/- 0.1 | 99.7 +/- 0.1 | 0.047 +/- 0.003 | 0.084 +/- 0.000 |
| Fraction aromatic | 99.7 +/- 0.1 | 99.5 +/- 0.1 | 0.060 +/- 0.002 | 0.109 +/- 0.001 |
| **Composition** | | | | |
| Fraction carbon | 98.6 +/- 0.2 | 97.0 +/- 0.0 | 0.061 +/- 0.003 | 0.106 +/- 0.000 |
| Fraction nitrogen | 99.6 +/- 0.2 | 96.1 +/- 0.1 | 0.097 +/- 0.004 | 0.132 +/- 0.000 |
| Fraction oxygen | 99.4 +/- 0.1 | 99.4 +/- 0.1 | 0.050 +/- 0.003 | 0.058 +/- 0.001 |

We analysed the novelty of molecules for large sets of molecules to measure the ability of the explorative nature of the models to cover the vastness of chemical space. For all datasets, the novelty decreases slowly over time. The novelty rates after 2M compounds are 66.3, 73.3% and 75.4% for architectures A (LSTM-LSTM 256/256), B (biLSTM-biLSTM 256/256) and C (biLSTM-biLSTM 256/256 1/4), respectively. We observed further that the novelty rate for architecture A was systematically lower than the degree of novelty of both bidirectional architectures B and C. These results suggest that the use of bidirectional LSTM units is beneficial to maintain a high degree of generativity for the trained model. We also evaluated the total number of generated molecules over time (figure 4B). The performance of the model have been expressed as efficiency which is the percentage of valid unique molecules. After 2M, architecture A with two consecutive LSTM-LSTM produced 1,470,543 unique molecules with an efficiency of 73.5%. Architectures B and C with bidirectional embedding and encoding layers have generated 1,566,535 (78.5% efficiency) and 1,602,018

(80.1% efficiency) unique molecules respectively. The use of bidirectional layers was thus highly beneficial to improve the efficiency of the generation process.

As expected, the novelty decreased over time as a result of saturation of the chemical space (figure 5). The novelty rate of smaller molecules with atom count 5-14 decreased strongly to moderately over time, while the novelty rate of larger molecules with atom count 15-24 increased moderately to strongly with time (figure 5). The novelty of the generated molecules gradually moved from smaller to larger molecules over time. The saturation thus shifted slowly from smaller to larger molecules during the generation process. Additionally, the results for the novelty rate showed a subtle shift when moving from architecture A to architectures B and C. These results were in line with the expectation that bidirectional layers were better reproduced the the size-related properties.

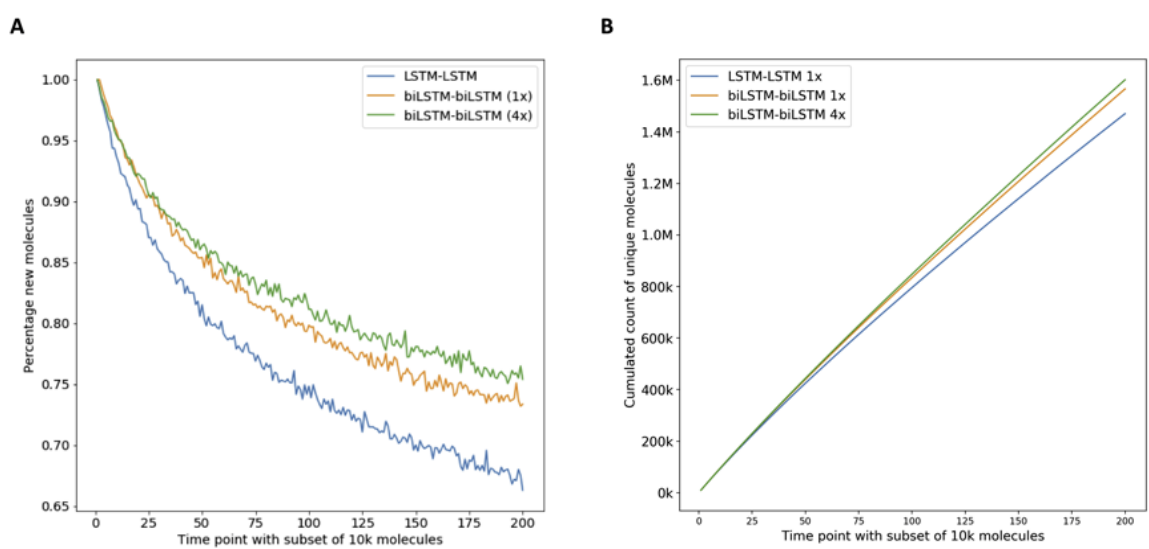

*Figure 4: **Global novelty analysis.** For all sets of 2M generated compounds, the dataset has been split into 10k time points. A) Plot showing the percentage of molecules at every time point t. B) Cumulated number of unique molecules generated during the process. The final values for the three tested architectures are 1,470,543 (73.5% efficiency) for LSTM-LSTM, 1,566,535 (78.3% efficiency) for biLSTM-biLSTM and 1,602,018 (80.1% efficiency) for biLSTM-biLSTM with 4 parallel concatenated encoding layers.*

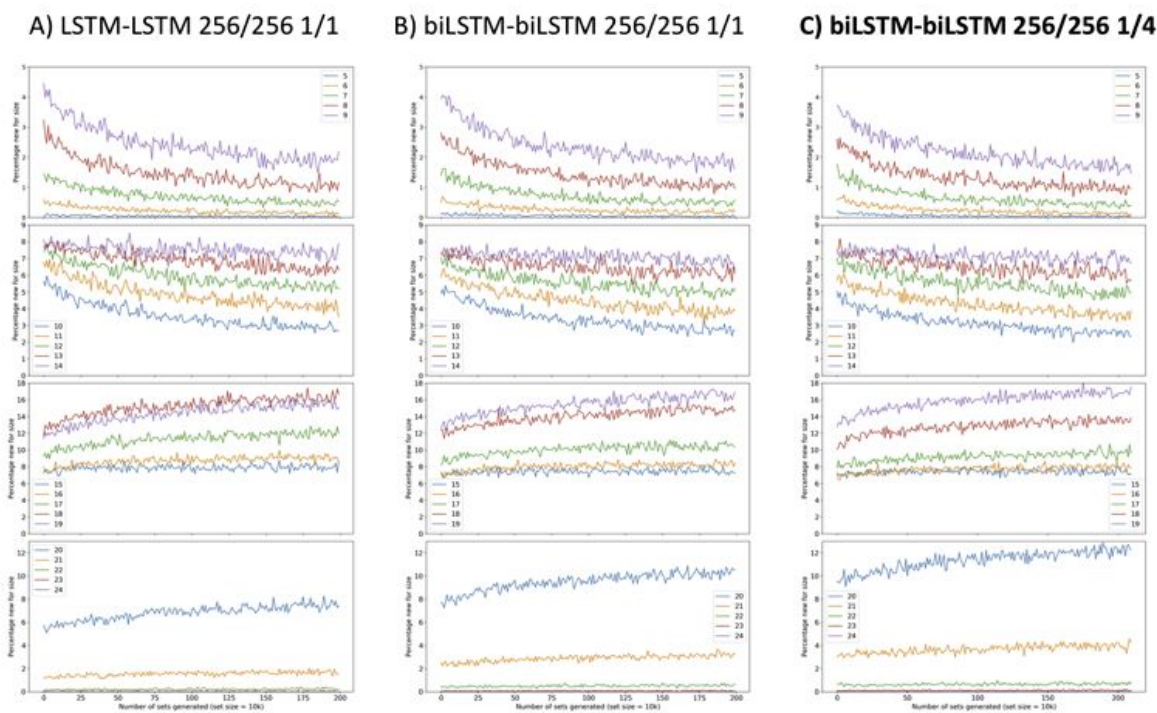

*Figure 5: Novelty analysis by atom count. the winning architecture is highlighted in bold (see methods).*

**Table 6: Impact of the training set size on GENs performance.**

| Dataset and size of subset | Augmented size (real factor[1]) | Best model epoch # | Validity % | Uniqueness % | Training % | Length match %[2] | HAC match %[3] |
|---|---|---|---|---|---|---|---|
| **PubChem 225k** | | | | | | | |
| 9k | 54,624 (4.8) | 10, 10, 10 | 81.3 +/- 0.9 | 100.0 +/- 0.0 | 0.3 +/- 0.1 | 97.7 +/- 0.0 | 90.5 +/- 0.0 |
| 45k | 218,124 (4.8) | 5, 5, 5 | 95.6 +/- 0.7 | 99.9 +/- 0.1 | 2.6 +/- 0.5 | 99.0 +/- 0.0 | 94.7 +/- 0.0 |
| 225k | 1,088,864 (4.8) | 4, 4, 4 | 98.3 +/- 0.3 | 99.9 +/- 0.0 | 11.2 +/- 0.5 | 97.3 +/- 0.7 | 96.6 +/- 0.3 |
| **Chembl24** | | | | | | | |
| 9k | 35,928 (4.0) | 44, 43, 45 | 74.2 +/- 1.9 | 99.0 +/- 0.2 | 0.2 +/- 0.2 | 81.9 +/- 5.4 | 95.9 +/- 1.0 |
| 45k | 179,888 (4.0) | 5, 6, 5 | 91.9 +/- 1.9 | 100.0 +/- 0.0 | 0.2 +/- 0.1 | 90.6 +/- 2.8 | 97.6 +/- 1.4 |
| 225k | 896,214 (4.0) | 9, 6, 6 | 94.6 +/- 0.1 | 100.0 +/- 0.0 | 1.4 +/- 0.3 | 88.4 +/- 1.6 | 98.1 +/- 0.6 |
| **Zinc15** | | | | | | | |
| 9k | 32,546 (3.6) | 24, 21, 21 | 77.2 +/- 1.0 | 100.0 +/- 0.0 | 0.0 +/- 0.0 | 82.2 +/- 3.3 | 91.2 +/- 1.1 |
| 45k | 163,929 (3.6) | 10, 7, 11 | 90.4 +/- 1.1 | 100.0 +/- 0.0 | 0.1 +/- 0.1 | 87.6 +/- 1.2 | 92.6 +/- 1.1 |
| 225k | 820,747 (3.6) | 4, 6, 6 | 95.2 +/- 0.3 | 100.0 +/- 0.0 | 0.3 +/- 0.1 | 90.4 +/- 1.2 | 93.5 +/- 1.2 |

1) Size of the augmented dataset after 5 random attempts per SMILES and deduplication to unique SMILES. Real augmentation factor varies depending on dataset.
2) Length match for SMILES length distributions of the training set and generated set (see methods).
3) HAC match for the atom count distributions of the generated set and training set (see methods).

Finally, we also measured the impact of the training set size on the performance of the GENs. This evaluation was done to investigate whether a well-defined architecture could autonomously learn the alphabet and grammar of the SMILES strings without the need for didactic feedback from either a discriminator or through reinforcement learning. In this comparison we tested models developed with PubChem, Chembl24 and Zinc15 datasets using 9k, 45k and 225k unique molecules, respectively (table 6). The results for all datasets showed clear improvements for the number of valid molecules and moderate improvements for the percentage match of heavy atom count with increasing the training set size. Indeed, the percentage of valid molecules for PubChem increased from 81.3 to 98.3, for Chembl24 from 74.2 to 94.6% and for Zinc from 77.2 to 95.2%. The small differences in numbers was expected and could be explained by the fact that we used a focused library for the PubChem dataset while training sets for the Chembl and Zinc were selected by chance. Other evaluation measures are relatively stable. These results are in line with the expectation that increasing the training set size is highly beneficial for learning the SMILES alphabet and grammar of the analysed data. The use of the large sets is shown to decrease the number of epochs required for the convergence of the generator. Examples of generated molecules for the models trained with Chembl24 and Zinc15 are displayed in figure 7A and 7B, respectively. The small set of selected examples are highlighting that the autonomously learning generator can easily handle complex SMILES and generates SMILES with a vivid curiosity and open-mindedness. Consequently, the generators are well equipped to explore new areas of chemical space.

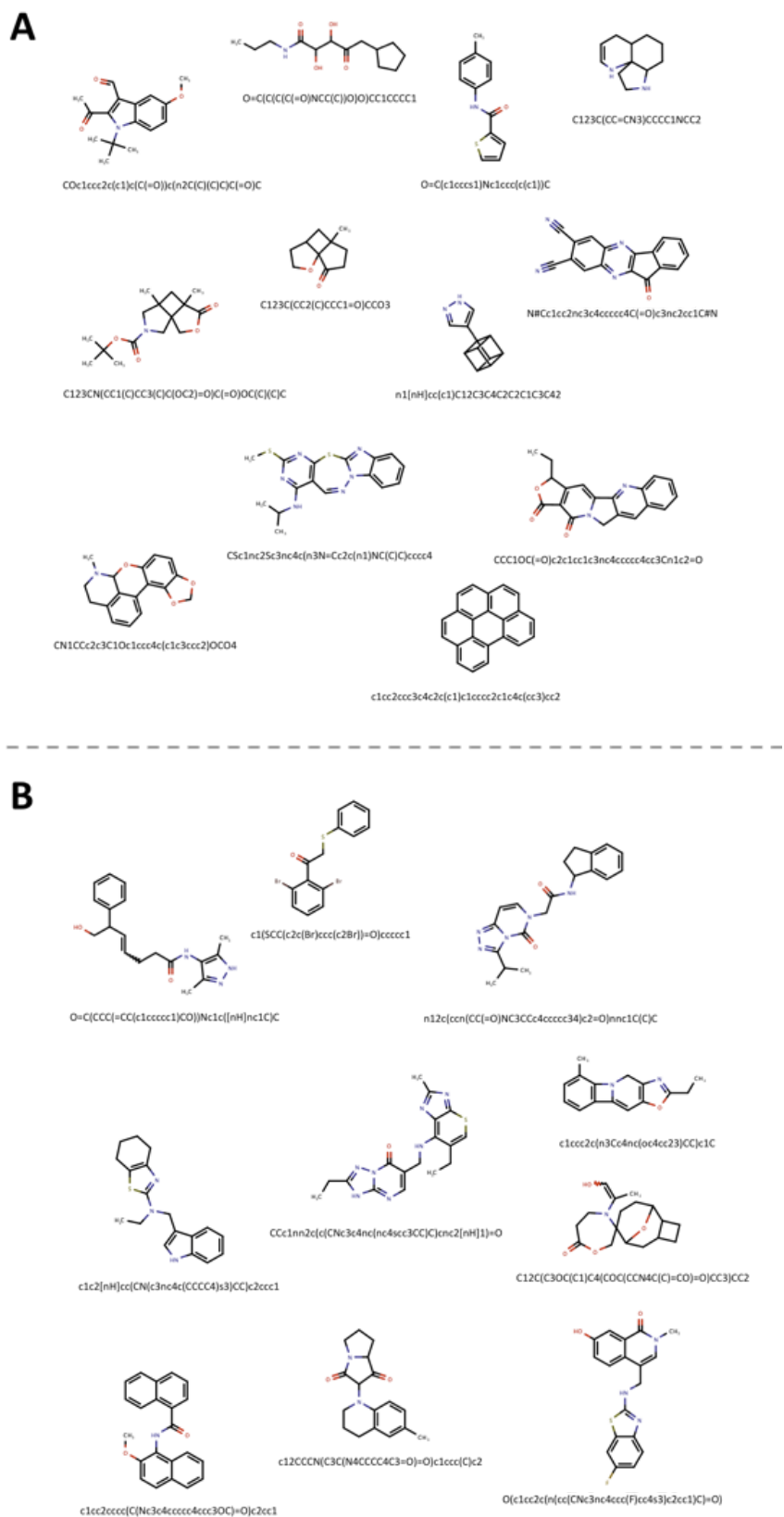

*Figure 6: Generated SMILES strings for Selected examples from Chembl24 and Zinc evaluated in table 6.* A) 12 molecules generated by the model after training with 225k randomly selected Chembl24 molecules. B) 10 selected examples of SMILES generated after training the model with 225k randomly selected molecules from Zinc15.

**Discussion.**

Our goal was to obtain a powerful SMILES based molecular generator which can generate a high degree of valid novel molecules while being within the property space of the training set. To obtain such results we fine-tuned an earlier published architecture [6]. Key modification is the introduction of an EarlyStopping mechanism with an online generator used to perform a statistical quality control at every epoch. This new feature allowed us to better monitor the learning process of the generator. The primary aim of this mechanism was to select the earliest stable model capable of generating a very high percentage of valid molecules as a method of overfit control [31]. Indeed, within a few training epochs, we reached very high percentages of valid SMILES produced by the generator (95.9-98.3%). Our analysis showed that the decision to stop training early based on the percentage of valid molecules, did not affect the capability of the model to generate SMILES with a high degree of novelty. The use of this EarlyStopping-mechanism may also be beneficial for use with RNN-based predictive models to partially freeze layers from further optimization.

Selection of LSTM in an architecture with parallel encoding layers, allowed us to reduce the number of hyperparameters in the network, while maintaining a stable generator with excellent generation results. We also explored the merging of the parallel layers using concatenation, averaging or by learnable average. The results were clearly in favor of concatenation. We also evaluated the use of parallel embedding-encoding layers (architecture D), but the results did not improve compared to architecture C with only one embedding layer. These results suggest that a single embedding is sufficient to adequately learn the training set SMILES.

Using the best architecture, we tested the effect of SMILES augmentation to further improve the ability of the model to generate a higher percentage of valid molecules and/or better reconstruct the property space. Our results demonstrated that augmentation increased the percentage of valid molecules from 97.0 to 98.3%. The models developed using high augmentation provided nearly perfect reconstruction of the property space (97.1% to 98.2%). However, these models had a lower novelty of generated molecules. The introduced GENs generated SMILES string with comparable or even better quality to the recently published RNN-based SMILES generator [32,33,34,35]. Contrary to the earlier work,

GENs reached these results using a significantly smaller training set and fewer training iterations, due to the introduction of the online statistical quality control.

**GAN comparison.**

In a typical GAN architecture two networks, i.e. a Generator and a Discriminant, compete one with another. To our knowledge the closest method to our GEN is SeqGAN [36]. SeqGAN is modeling the generator as a stochastic policy. The reward signal coming from the discriminator is judged on a complete sequence, and gradient is passed back to the intermediate state-action steps using Monte Carlo search. Recent introduction of wasserstein distance in GAN (WGAN) improves the generator [37,38] by using a smooth metric for measuring the distance between two probability distributions.

In GENs, a discriminator is absent and is replaced by an independent examiner, which runs a statistical assessment of the generator output after every epoch. GANs typically need full datasets to perform a sound evaluation. The proposed in this work method has only one neural network to convert SMILES sequences into a probabilistic model for prediction of the next character in a SMILES string. Its generator mecanism is autonomously learning the training set and it is not influenced by the feedback from the examiner. This differentiate GENs from GANs or models with reinforcement learning (RL), which both require a feedback mechanism. Nevertheless, as demonstrated in this work, GENs achieve spectacular results on the reconstruction of the chemical space of the training set with a vivid curiosity and open-mindedness. The latter is expected to be the result of the GEN methodology allowing the generator to acquire the knowledge by self-directed learning while being independently examined.

GENs are open setup with an early-stopping mechanism. Training can be easily continued and GENs are thus could be potentially used in different applications such as transfer learning (TL) [39].

If we treat SMILES as text, one can notice that SMILES contain two major graph conversion challenges, ring and branch representation in 1D. They can be considered as analogs of grammar and conjugation in natural languages. However, SMILES only contain a small amount of unique characters, i.e., chemical "words". Based on the excellent results we observed using GENs, we believe that this limiting number of words, can be deciphered very quickly by neural networks when selecting a right architecture. Moreover, the overtrain of the deep generative models will even lead to the loss of the novelty of generated structures. The latter problem is also a classical issue known to GANs. In our opinion, a generator should learn the domain space but at the same time it must also have sufficient freedom to apply the extracted rules and maintain diversity of generated answers. The proposed examination of the quality of the learned chemical space can be used with other types of generators. Of course, the optimal examination mechanism will need to be defined and fine-tuned on a case-by-case basis. It is important to highlight, that the independent quality mechanism introduced in this work does not influence the generator, which is just autodidactic. In summary, the introduced GENs are along the recent development of artificial intelligence theory. A GEN can learn by itself and its ability to generalise the knowledge are checked by a quality test, i.e., in a similar way as IBM's Watson has recently passed the physician exams, as any other Homo Sapiens student of a college [40], to prove its knowledge.

**Conclusion.** The main goal of a generator is to produce a set of SMILES with a high degree of novelty while staying focused on the property space of the training set. By small adjustments to the existing architecture we obtained remarkable results for both these goals. Our GENs autonomously learn the alphabet and grammar of SMILES string to generate valid molecules within the target space within a few epochs. The winning architecture uses an ensemble of smaller networks, capable of achieving similar results as larger networks [41]. The introduced early-stopping mechanism of the generators allows to maintain a high degree of novelty thanks to online statistical quality control, which measures the percentage of valid SMILES. The EarlyStopping mechanism is easily adaptable and open to accommodate other quality metrics such as distribution overlap, or multi-objective targets or other models. GENs are thus can be used for transfer learning or domain adaptation tasks. After the EarlyStopping, training of the GENs can be continued to tackle new challenges [39]. The analysis of different architectures showed that the use of a bidirectional embedding layer followed by multiple parallel encoding layers was essential for stable and fast training of the generator. The SMILES augmentation increased the volume of the training set and accuracy of produced models without the need for a larger set of diverse molecules. The augmentation improved stability of the generator to generate molecules. The code including example notebooks is distributed freely under a Clause-3 BSD License (https://opensource.org/licenses/BSD-3-Clause) [21].

**Author Contributions**

PE provided the initial architecture. RvD has introduced the methods for live monitoring and online statistical quality control. This architecture has been subsequently modified by RvD, IT and GG to include the bidirectional architecture of the embedding and encoding layers including more advanced architectures with concatenated and averaged layers. RvD applied the calculations for hyperparameter optimization and creation of the models. GG and RvD applied the calculations for the generation and analysis of the generated molecules. All authors contributed to the writing of the manuscript.

**Corresponding authors**

Guillaume Godin and Ruud van Deursen, Firmenich SA, Research and Development, Rue des Jeunes 1, CH-1227 Les Acacias, Geneva, Switzerland, guillaume.godin@firmenich.com, ruud.van.deursen@firmenich.com

**Competing interests**
The authors declare that they do not have any competing interests.